\documentclass[sigconf]{acmart}

\AtBeginDocument{%
  }

\copyrightyear{2026}
\acmYear{2026}
\setcopyright{cc}
\setcctype{by-nc-nd}
\acmConference[MM '26]{Proceedings of the 34th ACM International Conference on Multimedia}{November 10--14, 2026}{Rio de Janeiro, Brazil}
\acmBooktitle{Proceedings of the 34th ACM International Conference on Multimedia (MM '26), November 10--14, 2026, Rio de Janeiro, Brazil}
\acmDOI{10.1145/3767308.3835086}
\acmISBN{979-8-4007-2213-4/2026/11}

\usepackage{caption}
\usepackage{multirow} 
\usepackage{pifont}

\usepackage{caption}
\usepackage{subcaption}
\usepackage{multirow}  
\usepackage{amsthm}

\usepackage[table]{xcolor} 

\definecolor{LightBlue}{RGB}{240,248,255}
\definecolor{deepgray}{gray}{0.3}

\newcommand{\cmark}{\ding{51}}

\newcommand{\app}{\raise.17ex\hbox{$\scriptstyle\sim$}} 
\definecolor{baselinecolor}{gray}{.9}

\newcolumntype{^}{>{\currentrowstyle}}

\definecolor{dt}{gray}{0.7}  %

\usepackage[capitalize]{cleveref}
\crefname{section}{Sec.}{Secs.}
\Crefname{section}{Section}{Sections}
\Crefname{table}{Table}{Tables}
\crefname{table}{Tab.}{Tabs.}

\newcolumntype{S}{@{}>{\lrbox0}l<{\endlrbox}}  %
\definecolor{lightgreen}{HTML}{D8ECD1}

\newcommand{\ie}{{\emph{i.e.}}, }

\newcommand{\eg}{{\emph{e.g.}}, }
 
\begin{document}
 
\title{Bridging the Gap Between Semantics and Reconstruction:
Unifying Sign Language Translation and Production}


\author{Xiao Liu}
\orcid{0000-0001-6943-9861} 
\affiliation{%
  \institution{State Key Laboratory of Novel Software Technology, Nanjing University}
  \city{Suzhou}
  \country{Jiangsu}
    \country{China}
}
\email{liuxiaox@smail.nju.edu.cn}

\author{Shiwei Gan}
\authornote{Corresponding authors.} 
\orcid{0000-0003-3360-4321} 
\affiliation{%
  \institution{State Key Laboratory of Novel Software Technology, Nanjing University}
  \city{Nanjing}
  \state{Jiangsu}
  \country{China}
  }
\email{sw@nju.edu.cn} 

\author{Yafeng Yin}
\orcid{0000-0002-9497-6244} 
\authornotemark[1]
% \authornote{Corresponding author.} 
\affiliation{
  \institution{State Key Laboratory of Novel Software Technology, Nanjing University}
  \city{Suzhou}
  \country{Jiangsu}
    \country{China}
}
\email{ yafeng@nju.edu.cn}

\author{Jiaxin Yin}
\orcid{}  
\affiliation{
  \institution{State Key Laboratory of Novel Software Technology, Nanjing University}
  \city{Suzhou}
  \country{Jiangsu}
    \country{China}
}
\email{jiaxin.yin@smail.nju.edu.cn}

\author{Bowen Guo}
\orcid{0009-0000-6390-4398} 
\affiliation{
  \institution{State Key Laboratory of Novel Software Technology, Nanjing University}
  \city{Suzhou}
  \country{Jiangsu}
    \country{China}
} 
\email{bowen@smail.nju.edu.cn}

\author{Yaqi Sun}
\orcid{0009-0007-9520-9745} 
\affiliation{
  \institution{State Key Laboratory of Novel Software Technology, Nanjing University}
  \city{Suzhou}
  \country{Jiangsu}
    \country{China}
}  
\email{yaqi@smail.nju.edu.cn}

\author{Zhiwei Jiang}
\orcid{0000-0001-5243-4992}  
\affiliation{
  \institution{State Key Laboratory of Novel Software Technology, Nanjing University}
  \city{Suzhou}
  \country{Jiangsu}
    \country{China}
} 
\email{jzw@nju.edu.cn}

\author{Lei Xie}
\orcid{0000-0002-2994-6743} 
\affiliation{%
  \institution{State Key Laboratory of Novel Software Technology, Nanjing University}
  \city{Nanjing}
  \state{Jiangsu}
  \country{China}
} 
\email{lxie@nju.edu.cn}

\renewcommand{\shortauthors}{Xiao Liu et al.}

\begin{abstract}
\input{Section/abstract}
\end{abstract}




\begin{abstract}
Recent advances in sign language (SL) research have shown a trend toward unifying multiple sign language understanding (SLU) subtasks, such as isolated sign language recognition (ISLR), continuous sign language recognition (CSLR), and sign language translation (SLT), within a single framework, leading to substantial progress. Meanwhile, sign language production (SLP), which generates sign sequences from text, has also attracted growing attention. This naturally raises an important question: can sign language understanding and production be unified within a single framework?
Compared with unifying SLU subtasks, this problem is substantially more challenging. Existing SLU tasks largely share the same direction of mapping, namely from sign inputs to linguistic outputs, whereas SLT and SLP lie in opposite directions of sign-text mapping. A unified framework must therefore address two key challenges: (1) bridging the modality gap between continuous sign motions and discrete text tokens through a shared sign tokenizer that supports both linguistic abstraction and motion reconstruction; and (2) learning a single conditional autoregressive model that can take either sign or text as input and generate the corresponding target sequence in the opposite modality.
To this end, we propose Uni-SLTP, a unified framework for SLT and SLP with two key components: (1) a shared sign tokenizer that converts sign sequences into discrete tokens and latent representations, capturing both semantic and reconstructive information; and (2) a unified autoregressive generation model that formulates both tasks as conditional sequence generation. Experiments on widely used public datasets show that Uni-SLTP achieves superior motion accuracy for SLP while maintaining competitive SLT performance.
\end{abstract}

\maketitle

\section{Introduction} 
\label{sec:introduction}
Sign language (SL) is the primary means of communication for the hearing-impaired  community. Existing research primarily focuses on two directions: sign language understanding (SLU)~\cite{li2025uni,zuo2023natural}, with sign language translation (SLT) receiving the most attention~\cite{gueuwou2025signmusketeers,zhou2023gloss,ye2024improving,liang2024llava,gong2024llms} and sign language production (SLP)~\cite{zuo2025signs,baltatzis2024neural,stoll2022there,yin2023gloss,tang2025gloss}. SLT aims to convert SL sequence into natural language text, making it easier for the general public to understand, while SLP seeks to translate natural language text into SL sequences, enabling the hard of hearing individuals to better access and understand information. 

\begin{figure}[t]
  \centering
    \includegraphics[width=1.0\columnwidth]{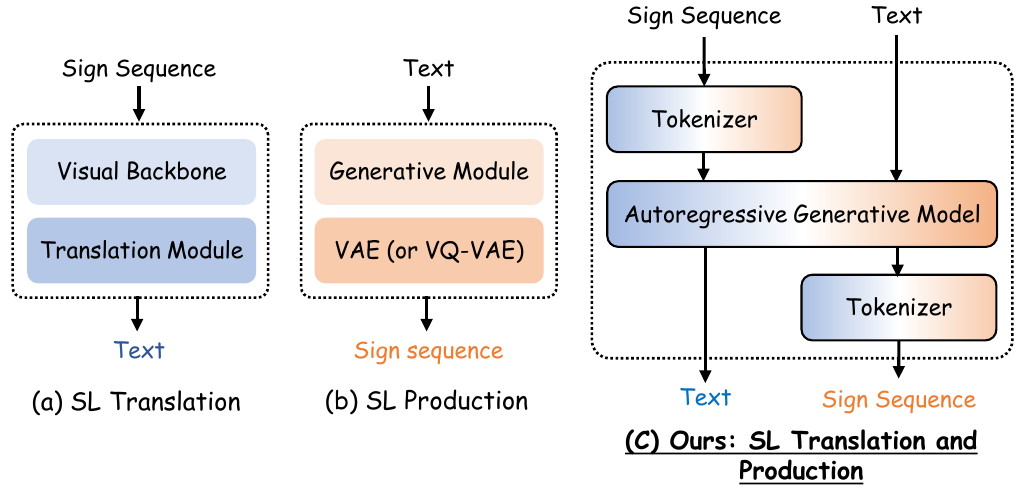}
    \vspace{-3mm}
   \caption{Previous task-specific SLT/SLP models vs Uni-SLTP.}
   \label{fig:motivation} 
  \vspace{-3mm}
\end{figure}

Considering that the two tasks involve fundamentally different modality transformations, SLT maps continuous sign sequences to discrete text tokens, whereas SLP converts discrete text sequences into continuous, natural sign sequences, existing research typically models them with separate architectures. As shown in Figure~\ref{fig:motivation} (a), for SLT, the de facto architecture typically begins with a visual backbone that extracts sign semantic features across both spatial and temporal dimensions, followed by a translation model (\eg mBART~\cite{liu2020multilingual} or GPT-2~\cite{ethayarajh2019contextual}), to generate the corresponding text sentence.
For SLP tasks shown in Figure~\ref{fig:motivation} (b), current methods mainly follow two paradigms: (1) Diffusion-based approaches~\cite{baltatzis2024neural,qi2024signgen}: 
A text-conditioned diffusion model generates continuous or discrete latent features learned by a VAE or VQ-VAE, which are then decoded into target sign sequences.
(2) Autoregressive (AR)-based methods~\cite{yin2024t2s,ma2024ms2sl,zuo2025signs}: A VQ-VAE converts the sign sequence into discrete IDs, allowing SLP to be modeled as text-conditioned sequence prediction. An AR model predicts these IDs, which the VQ-VAE decodes into the target sign sequence.

Meanwhile, recent advances in SL understanding (SLU), such as Unisign~\cite{li2025uni}, BEST~\cite{zhao2023best}, MSLU~\cite{zhou2025scaling}, have explored unifying multiple SLU tasks, including isolated SL recognition (ISLR)~\cite{zuo2023natural,hu2021hand, li2020transferring}, continuous SL recognition (CSLR)~\cite{gan2024signgraph, wei2023improving, hu2023self}, and SLT, within a single framework. 
% These models significantly reduce the modeling complexity of individual tasks through a unified paradigm.
These models significantly reduce individual-task modeling complexity through a unified paradigm.
\textbf{Beyond SLU, a more fundamental question arises: Can SLU (\eg SLT) and SLP be unified within a single framework to enable shared token representations and bidirectional generation}? 
Our answer is that unifying these tasks is feasible, but not straightforward. (1) For  unified SLU models~\cite{zhou2025scaling,zhao2023best}, 
although different understanding tasks may have different output forms, they generally share the same mapping direction, \ie from sign inputs to linguistic outputs such as glosses or text sequences.
% which focus on unifying understanding tasks, both the input and output of these models remain within the same modality. Specifically, they take sign sequences as input and produce textual output (\eg glosses, text sequences).  
(2) In the unified SLU settings, the intermediate sign representations do not introduce conflicting objectives. All subtasks benefit from learning stronger semantic representations of SL, making the optimization of the shared sign backbone relatively straightforward.

%, which we refer to as the \textbf{SLTP} task,
Unlike previous unified SLU models, unifying SLT and SLP is more challenging, considering: \textbf{ (1) The two tasks differ fundamentally in their input and output modalities.}  Specifically, SLT maps sign sequences to text, whereas SLP maps text to sign sequences, making it non-trivial to formulate a single unified task for joint modeling. 
% A natural idea, inspired by recent advances in image understanding and generation~\cite{wu2024vila,ma2025unitok,qu2025tokenflow,xie2025muse}, 
A possible approach is to introduce a sign tokenizer using VQ-VAE (referred to as a \textit{sign tokenizer}), which converts sign sequences into discrete intermediate representations. 
The relationship between these discrete sign representations and discrete text tokens can then be modeled in a bidirectional autoregressive manner;
\textbf{(2) However, simply adopting the same sign tokenizer to simultaneously capture low-level motion details required for SLP and high-level semantic representations required for SLT is inherently difficult.}
SLT and SLP impose conflicting requirements on sign representations: translation benefits from semantic abstraction and alignment with text, while production demands fine-grained motion accuracy to enable natural and accurate synthesis. 
We term this tension the \textit{Semantic--Reconstruction Gap}, which makes it hard to learn discrete sign tokens that are both semantically aligned and highly decodable for generation.

% This inherent tension, which we term the \textit{Semantic--Reconstruction Gap}, makes it challenging to learn discrete sign tokens that are simultaneously well aligned with text semantics and highly decodable for sign generation.

In this paper, we revisit SLT and SLP and aim to develop a general unified framework supporting both tasks with two main objectives. 
(1) A unified autoregressive model that operates on both sign and text representations, and can flexibly generate either text or sign token sequences, thereby enabling joint modeling of SLT and SLP within a single framework.
(2) A unified sign tokenizer that provides effective intermediate representations of sign language, capturing both fine-grained motion details for accurate reconstruction and high-level semantic information for sign understanding. 

To achieve these goals, we propose \textbf{Uni-SLTP}, a unified framework that supports bidirectional mapping between sign and text within a single architecture. 
Specifically, to build a shared sign tokenizer that serves both SLT and SLP tasks, providing discrete representations that capture both fine-grained motion details for reconstruction and high-level semantic features for understanding, we introduce \textbf{Semantic-Reconstruction guided Residual Vector Quantization (SR-RVQ)} as a shared discrete interface that decouples semantic alignment from detail reconstruction.  
To allow modeling  SLT and SLP in the same framework,  we cast both SLT and SLP as conditional next-token prediction in a unified pipeline via proposed SR-RVQ and AR modeling, enabling bidirectional learning with competitive semantic consistency or motion accuracy.

\begin{itemize}
    \item To the best of our knowledge, we are the first to formulate SLT and SLP as bidirectional sign--text generation. We propose \textbf{Uni-SLTP}, a unified framework that casts both directions as conditional next-token prediction in an autoregressive pipeline, enabling one model to perform both tasks.
    
    \item We introduce \textbf{Semantic-Reconstruction guided Residual Vector Quantization (SR-RVQ)}, a hierarchical tokenizer that decouples semantic alignment from motion-detail reconstruction via a semantic anchor token and residual motion tokens, bridging the Semantic--Reconstruction Gap.

    \item Extensive experiments show that Uni-SLTP improves SLP quality while maintaining competitive SLT performance against task-specific baselines. 
\end{itemize}

\section{Related Work}
\paragraph{Sign Language Translation.} 
SLT aims to translate sign language (SL) sequences into textual sentences.~\nocite{wei2020semantic,pu2018dilated,cheng2020fully,zuo2022c2slr,koller2019weakly,pu2020boosting,hao2021self,min2021visual,zhang2019continuous,koller2017re,huang2018video,tang2021graph,guo2018,camgoz2017subunets,yang2019sf,cui2017recurrent,niu2020stochastic,zhang2023c2st,hu2022temporal,zhu2024multiscale,wei2023improving, hu2023self,parelli2022spatio,jiao2023cosign,hu2023continuous,wei2023improving,zuo2022c2slr,koller2019weakly,koller2017re,tang2021graph,gan2024signgraph,niu2020stochastic,zhou2020spatial,zhou2019dynamic,cui2019deep} Existing approaches typically employ a sign tokenizer (\ie a visual backbone) to encode SL sequences into intermediate sign representations, which are then fed into a pretrained translation model (typically an autoregressive (AR) language model like mBART~\cite{chen2022two}, GPT-2~\cite{gong2024llms}) to generate the target text. Despite the strong language modeling capabilities of language models (LMs), recent studies indicate that the primary performance bottleneck of SLT models lies in extracting effective and semantically accurate sign representations.
Consequently, most prior work adopts a two-stage training paradigm and proposes various strategies to pretrain sign tokenizers, including CTC-based gloss supervision~\cite{zhou2021improving, chen2022simple}, contrastive learning with text~\cite{liang2024llava, jiao2024visual}, pseudo-gloss supervision~\cite{gan2025mixsigngraph, guo2025bridging}, and large-scale SL dataset pretraining~\cite{gueuwouetal2025shubert, li2025uni, zhao2023best}.
Such pretraining strategies enable the sign tokenizer to learn more effective and sign-specific semantic representations, which in turn provide high-quality inputs to the translation module, and are crucial for ensuring translation performance.

% \vspace{-3mm}
\paragraph{Sign Language Production.} Sign language production (SLP) \cite{liu2026signpr,saunders2020adversarial,saunders2020progressive,saunders2021mixed,tang2025gloss,xie2024g2p,zuo2024simple,rastgoo2021sign} aims to generate sign sequences conditioned on spoken-language text.~\nocite{fang2023signdiff,baltatzis2024neural,ma2024ms2sl,rastgoo2021sign,saunders2020adversarial,saunders2020progressive,saunders2021continuous,saunders2021mixed,saunders2022signing,stoll2018sign,stoll2022there,tang2025gloss,walsh2024data,yin2024t2s,tang2025sign,ma2024ms2sl,zuo2025signs,walsh2024data} To avoid the high dimensionality and redundancy of raw videos, recent methods increasingly adopt pose sequences or parametric body trajectories as the generation target \cite{saunders2022signing,fang2023signdiff}. Existing SLP approaches can be broadly grouped by whether they rely on gloss as an intermediate supervision signal. Gloss-based pipelines \cite{stoll2018sign,stoll2022there,yin2023gloss,tang2025gloss} leverage glosses to facilitate learning and enforce monotonic alignment, but they depend on costly and unevenly available gloss annotations. This has motivated a growing line of gloss-free, end-to-end text-to-pose generation \cite{yin2024t2s,ma2024ms2sl,baltatzis2024neural}, where duration and temporal alignment are modeled either explicitly or implicitly within sequence generators. Early end-to-end text-to-pose methods largely adopt continuous regression-based encoder–decoder generators \cite{saunders2020adversarial,saunders2021continuous}. More recently, two paradigms have attracted attention: diffusion-based methods that synthesize pose sequences via iterative denoising under text conditioning \cite{baltatzis2024neural}, and discrete autoregressive approaches that tokenize poses and perform next-token prediction, enabling efficient decoding and easier integration with language models \cite{zuo2025signs}.

SLT and SLP have advanced with task-specific architectures, while SLU has begun to unify sign understanding objectives.
Yet SLT and SLP remain largely separated, as they map between continuous motion and discrete text in opposite directions, demanding language-aligned and faithfully decodable representations.
By discretizing signs into tokens, tokenizers provide a shared interface that narrows the modality gap between sign and text.
This makes it possible to cast both directions as conditional next-token prediction under a single autoregressive framework.
The remaining challenge is to learn a token space that reconciles semantic alignment for SLT with fine-grained reconstructability for SLP.

\begin{figure*}[t]
  \centering
    \includegraphics[width=1.0\linewidth]{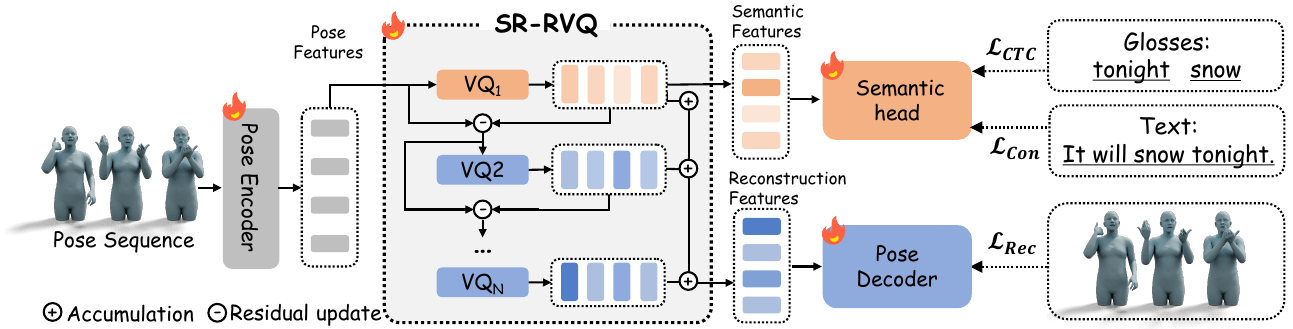}
    \vspace{-1mm}
   \caption{SR-RVQ tokenizer pretraining. A pose encoder maps sign pose sequences to latent features, which are quantized by SR-RVQ into semantic tokens $Q_1$ and residual detail tokens \(Q_{2:N}\). $Q_1$ serves as a semantic anchor trained with gloss CTC and semantic contrastive supervision, while \(Q_{2:N}\) are optimized with reconstruction loss to capture fine-grained motion details.}
   \label{fig:tokenizer} 
  \vspace{-2mm}
\end{figure*} 

\section{Method}

\subsection{Preliminaries}
\label{sec:prelim}
We first define the key modules used in our framework: \\
\noindent (1) \textbf{Sign Tokenizer ($\mathbf{ST}$):} a sign encoder $\mathbf{ST}_e$ that maps SL inputs into discrete  IDs and embeddings, and a sign decoder $\mathbf{ST}_d$ that reconstructs SL sequences from token IDs. \\
\noindent (2) \textbf{Text Tokenizer ($\mathbf{TT}$):} a text encoder $\mathbf{TT}_e$ that maps text inputs into token IDs and embeddings\footnote{Standard text tokenizers typically do not include embedding functions; we include them here for consistency and clarity.}, and a text decoder $\mathbf{TT}_d$ that converts token IDs back into text sequences.\\
\noindent (3) \textbf{Autoregressive Module ($\mathbf{AR}$):}  modeling the conditional generation of target token IDs in an AR manner.

\paragraph{SLT Task Formulation.} SLT  is formulated as a conditional sequence generation task, aiming to generate a target text sentence $\mathbf{W}$ from a continuous SL sequence $\mathbf{S}$. 
In practice, an SLT model first employs a sign tokenizer to map the input sequence into latent representations: 
$\mathbf{E_s} = \mathbf{ST}_e(\mathbf{S})$,
and then uses the autoregressive module $\mathbf{AR}$ to generate the target token sequence $\mathbf{W}^{id}$ conditioned on $\mathbf{E_s}$. 
The conditional distribution is factorized as
\begin{equation}
p_\theta(\mathbf{W}^{id} \mid \mathbf{E_s})
= \prod_{u=1}^{U} p_\theta\!\left(w_u^{id} \mid \mathbf{W}^{id}_{<u}, \mathbf{E_s}\right),
\end{equation}
where $\mathbf{W}^{id}_{<u}$ denotes the previously generated tokens. Finally, the text decoder reconstructs the output sentence from the token IDs:
$\mathbf{W} = \mathbf{TT}_d(\mathbf{W}^{id})$.

\paragraph{SLP Task Formulation.} SLP is commonly studied under either diffusion-based or autoregressive (AR)-based paradigms. To enable a unified formulation of SLP and SLT, we focus on the AR-based approach.   In this framework, an SLP model first employs a text tokenizer to encode a textual input sequence into latent representations:
$ \mathbf{E}_t = \mathbf{TT}_e(\mathbf{W})$, and then uses $\mathbf{AR}$ module to generate discrete SL tokens, where the conditional distribution is factorized as
\begin{equation}
p_\theta(\mathbf{S}^{id} \mid \mathbf{E}_t)
= \prod_{u=1}^{U} p_\theta\!\left(s_u^{id} \mid \mathbf{S}^{id}_{<u}, \mathbf{E}_t\right),
\end{equation}
Finally, the $\mathbf{ST}_d$ reconstructs the SL sequence from the predicted token IDs: $\mathbf{S} = \mathbf{ST}_d(\mathbf{S}^{id})$.

\paragraph{Unified Formulation of SLT and SLP.}  
From a unified perspective, both SLT and SLP can be formulated as AR sequence modeling problems over discrete tokens. Specifically, given an input sequence $X$ and a target sequence $Y$, 
a \emph{source} tokenizer $T=(T_e,T_d)$ encodes $X$ into source embeddings $\mathbf{E}=T_e(X)$,
and a \emph{target} tokenizer $D=(D_e,D_d)$ encodes $Y$ into target token IDs $\mathbf{Y}^{id}=D_e(Y)$.
The model is trained to learn the conditional distribution
\begin{equation}
p_\theta(\mathbf{Y}^{id} \mid \mathbf{E})
= \prod_{u=1}^{U} p_\theta(y^{id}_u \mid \mathbf{Y}^{id}_{<u}, \mathbf{E}),
\end{equation}  During inference, the AR predicted token sequence ${\mathbf{Y}}^{id}$ is decoded into the output sequence $\mathbf{Y}$ using the target tokenizer:
${\mathbf{Y}} = D_d({\mathbf{Y}}^{id})$.
Under this unified formulation, SLT and SLP thus share a single modeling paradigm, with the input and output swapped between the two directions. 
\begin{itemize}
    \item \textbf{SLT:} The input $X$ is a sign sequence $\mathbf{S}$, and the output $Y$ is a text sentence $\mathbf{W}$. We use $T=\mathbf{ST}$ to encode $\mathbf{S}$ into source embeddings, and $D=\mathbf{TT}$ to decode predicted IDs into $\mathbf{W}$.
    \item \textbf{SLP:} The input $X$ is a text sentence $\mathbf{W}$, and the output $Y$ is a sign sequence $\mathbf{S}$. We use $T=\mathbf{TT}$ to encode $\mathbf{W}$ into source embeddings, and $D=\mathbf{ST}$ to decode predicted IDs into $\mathbf{S}$.
\end{itemize}

In both SLT and SLP, we use the pretrained LM's built-in text tokenizer $\mathbf{TT}$ for text tokenization and decoding.
To unify the two directions, the remaining challenge is to (i) learn a discrete sign tokenization that converts continuous sign into a sequence of pose token IDs, and (ii) train a single AR backbone $\mathbf{AR}$ to model both text tokens and pose tokens under a shared vocabulary.
Next, we introduce our sign tokenizer and describe the unified AR modeling.

\vspace{-2mm}
\subsection{Semantic-Reconstruction Guided Sign Tokenizer}
\label{sec:tokenizer}
We introduce the Semantic-Reconstruction Guided Residual Vector Quantizer (SR-RVQ), a pose tokenizer that discretizes continuous sign motion into pose tokens for unified AR modeling, as shown in Figure~\ref{fig:tokenizer}.
 Our goal is to learn a discrete pose representation that simultaneously (i) aligns with text for SLT and (ii) preserves motion details for accurate reconstruction. This calls for tokens that are both \emph{text-aligned} and \emph{reconstruction-accurate}.
However, jointly enforcing semantic alignment and reconstruction with a single discrete bottleneck is inherently unstable, as the two objectives may conflict during optimization~\cite{qu2025tokenflow}.
 To satisfy both requirements, we adopt an $N$-stage residual vector quantizer (RVQ)~\cite{lee2022autoregressive} to build a coarse-to-fine hierarchy: the first stage produces an alignment-oriented semantic stream, while later stages encode residual motion details to progressively refine reconstruction.

\paragraph{Pose Encoder.}Inspired by previous SLT models~\cite{gan2025mixsigngraph}, our pose encoder adopts a similar design. The inputs are processed by a ResNet1D to extract spatial features, 
which are then processed by a temporal convolution module to capture short-term temporal dependencies and perform temporal downsampling. 
Finally, a BiLSTM is applied to model long-term temporal relationships. 
Specifically, given a SL pose sequence $\mathbf{S}\in\mathbb{R}^{T\times D}$, the encoder outputs continuous latent features
$\mathbf{Z}=\mathcal{E}(\mathbf{S})\in\mathbb{R}^{T'\times d}$,
where $d$ is the latent dimension and $T'=\lfloor T/s\rfloor$ is obtained by a fixed temporal downsampling factor $s$.
We intentionally bias the encoder toward semantics, since fine-grained motion details can be recovered by subsequent residual stages and the pose decoder $\mathbf{ST}_d$.

\begin{figure*}[t]
  \centering
    \includegraphics[width=1.0\linewidth]{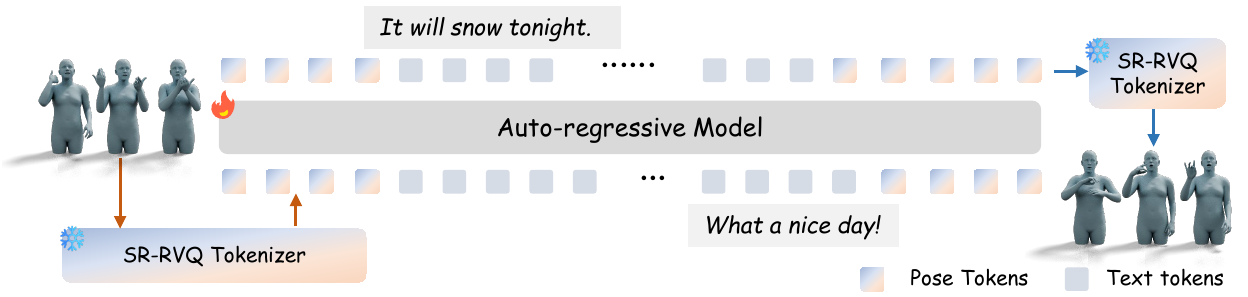}
    \vspace{-4mm}
   \caption{Uni-SLTP: unified autoregressive modeling for SLT and SLP. SR-RVQ converts pose sequences into discrete pose tokens, enabling a single AR model to perform conditional next-token prediction in both directions: pose-to-text for SLT and text-to-pose for SLP. } %Generated pose tokens are decoded back to continuous pose sequences by the pose decoder.
   \label{fig:framework} 
  \vspace{-2mm}
\end{figure*}

\paragraph{SR-RVQ.}
We discretize latents with an $N$-stage RVQ to obtain a coarse-to-fine hierarchy of token streams, where $N$ is the number of quantization stages.
Given latent features $\mathbf{Z}=\{\mathbf{z}_t\}_{t=1}^{T'}$, SR-RVQ produces $N$ index streams
$\{\mathbf{Q}_n\}_{n=1}^{N}$ with $\mathbf{Q}_n=\{q_{n,t}\}_{t=1}^{T'}$.
We maintain stage-wise token dictionaries $\mathcal{C}_n=\{\mathbf{c}^{(n)}_k\}_{k=1}^{K_n}\subset\mathbb{R}^{d}$, where $K_n$ is the size of the $n$-th dictionary.
% Given latent features $\mathbf{Z}=\{\mathbf{z}_t\}_{t=1}^{T'}$, where $\mathbf{z}_t\in\mathbb{R}^{d}$ denotes the latent feature at timestep $t$, SR-RVQ produces $N$ index streams $\{\mathbf{Q}_n\}_{n=1}^{N}$ with $\mathbf{Q}_n=\{q_{n,t}\}_{t=1}^{T'}$, where $q_{n,t}$ denotes the selected code index at stage $n$ and timestep $t$.
% We maintain stage-wise codebooks $\mathcal{C}_n=\{\mathbf{c}^{(n)}_k\}_{k=1}^{K_n}\subset\mathbb{R}^{d}$, where $\mathbf{c}^{(n)}_k$ denotes the $k$-th codeword in the $n$-th codebook and $K_n$ is its size.
At each timestep $t$, we initialize the residual $\mathbf{r}_{0,t}=\mathbf{z}_t$ and iteratively quantize:
\begin{equation}
\label{eq:rvq}
\begin{aligned}
q_{n,t} &= \arg\min_{k\in\{1,\dots,K_n\}}
\left\lVert \mathbf{r}_{n-1,t}-\mathbf{c}^{(n)}_k \right\rVert_2^2, \\
\mathbf{e}^{(n)}_t &= \mathbf{c}^{(n)}_{q_{n,t}}, \\
\mathbf{r}_{n,t} &= \mathbf{r}_{n-1,t}-\mathbf{e}^{(n)}_t ,
\end{aligned}
\end{equation}
where $\mathbf{r}_{n,t}$ denotes the stage-$n$ residual after subtracting the first $n$ token embeddings $\mathbf{e}^{(n)}_t$.
The quantized embedding is accumulated as $\hat{\mathbf{z}}_t=\sum_{n=1}^{N}\mathbf{e}^{(n)}_t$ (thus $\hat{\mathbf{Z}}=\{\hat{\mathbf{z}}_t\}_{t=1}^{T'}$).
In training, we assign semantic supervision primarily to the first stream $\mathbf{Q}_1$, while the remaining streams $\mathbf{Q}_{2:N}$ to recover residual motion details for reconstruction.

\paragraph{Pose Decoder.}
The pose decoder reconstructs the pose sequence from the quantized embeddings $\hat{\mathbf{Z}}$ to produce the reconstructed sequence $\hat{\mathbf{S}} = \mathbf{ST}_d(\hat{\mathbf{Z}})$. It begins with a 1D convolutional layer to process the latent embeddings, followed by two upsampling stages to recover the temporal details. Finally, a Conv1D projection layer maps the features back to the pose parameter space, producing the reconstructed pose sequence $\hat{\mathbf{S}} \in \mathbb{R}^{T \times D}$. 
% The detailed structure is illustrated in the Appendix.

\paragraph{SR-RVQ Optimization Objectives.}
We train the tokenizer with the principle: let the coarse stream capture high-level semantics, and let the residual streams fill in motion details.
This design leads to a two-part training objective: we (i) impose semantic supervision on the first-stage quantized features to encourage semantic alignment, and (ii) optimize reconstruction using the full quantized embeddings so that residual stages recover fine-grained motion details.
\textit{(1) Semantic-side.}
We apply semantic supervision only to the first-stage quantized features $\hat{\mathbf{Z}}^{(1)}=\{\hat{\mathbf{z}}^{(1)}_t\}_{t=1}^{T'}$ with $\hat{\mathbf{z}}^{(1)}_t=\mathbf{e}^{(1)}_t$ selected by $\mathbf{Q}_1$.
A lightweight CTC head on $\hat{\mathbf{Z}}^{(1)}$ predicts the gloss sequence $\mathbf{g}$, yielding $\mathcal{L}_{\text{ctc}}=-\log p(\mathbf{g}\mid \hat{\mathbf{Z}}^{(1)})$.
We further align pose semantics with the paired text sentence $\mathbf{y}$ using an InfoNCE loss. We compute a pooled pose embedding $\mathbf{h}_s=\mathrm{Pool}(\hat{\mathbf{Z}}^{(1)})$ and a pooled text embedding $\mathbf{h}_t=\mathrm{Pool}(\mathbf{TT}_e(\mathbf{y}))$, and use in-batch negatives:
\begin{equation}
    % \small
    \mathcal{L}_{\text{con}}
= - \frac{1}{B}\sum_{i=1}^{B}
\log \frac{\exp(\mathrm{sim}(\mathbf{h}_s^{(i)},\mathbf{h}_t^{(i)})/\tau)}
{\sum_{j=1}^{B}\exp(\mathrm{sim}(\mathbf{h}_s^{(i)},\mathbf{h}_t^{(j)})/\tau)} ,
\end{equation}
where $\mathrm{sim}$ is cosine similarity and $\tau$ is a temperature.
\textit{(2) Reconstruction-side.}
To recover motion details while keeping the first-stage codes semantics-oriented, we let the residual stages ($Q_{2:N}$) absorb most reconstruction pressure.
We optimize pose reconstruction with
$\mathcal{L}_{\text{rec}}
=\|\mathbf{S}-\hat{\mathbf{S}}\|_{1}
+\lambda_{v}\|\Delta\mathbf{S}-\Delta\hat{\mathbf{S}}\|_{1}$,
where $\Delta$ denotes the first-order temporal difference to encourage temporally coherent dynamics.
Concretely, we form the reconstruction latent as
$\hat{\mathbf{z}}^{\text{rec}}_t=\mathrm{sg}[\mathbf{e}^{(1)}_t]+\sum_{n=2}^{N}\mathbf{e}^{(n)}_t$,
so that gradients from $\mathcal{L}_{\text{rec}}$ do not update the stage-1 tokens, while residual stages learn to encode fine-grained motion variations.
Following standard vector quantization, we use the straight-through estimator and the commitment loss $\mathcal{L}_{\text{vq}}$ to learn the codebooks.
The overall tokenizer objective is
\begin{equation}
\label{eq:loss_tok}
\mathcal{L}_{\text{tok}}
=
\lambda_{\text{ctc}}\mathcal{L}_{\text{ctc}}
+\lambda_{\text{con}}\mathcal{L}_{\text{con}}
+\lambda_{\text{rec}}\mathcal{L}_{\text{rec}}
+\lambda_{\text{vq}}\mathcal{L}_{\text{vq}}.
% \vspace{-2mm}
\end{equation}

\subsection{Uni-SLTP: Unified Autoregressive Framework}
\label{sec:Uni-SLTP}
Uni-SLTP, as shown in Figure~\ref{fig:framework}, aims to support both SLT and SLP with a single  LM.
The key is to discretize continuous poses into SR-RVQ tokens and model text and pose tokens in one shared token space, so that both directions can be trained as conditional next-token prediction.

\paragraph{Unified vocabulary.}
We use the pretrained LM's built-in text tokenizer for spoken language, yielding text token IDs from the original vocabulary $\mathcal{V}_t$.
For poses, the frozen SR-RVQ tokenizer (Sec.~\ref{sec:tokenizer}) maps a pose sequence to stage-wise indices $\{q_{n,t}\}$.
To integrate pose tokens into the LM, we augment its vocabulary with a stage-aware pose vocabulary $\mathcal{V}_p$. For each RVQ stage $n$, we allocate a disjoint token block $\mathcal{V}_p^{(n)}$ of size $K_n$, together with boundary tokens such as \texttt{<sos>} and \texttt{<eos>}.
The unified vocabulary is $\mathcal{V}=\mathcal{V}_t\cup\mathcal{V}_p$. Accordingly, we extend both the LM embedding and output layers, where the parameters associated with $\mathcal{V}_t$ are inherited from the pretrained LM, while those associated with $\mathcal{V}_p$ are newly initialized and learned.
During decoding, we mask the output space to the valid subset: text decoding uses $\mathcal{V}_t$, while pose decoding at stage $n$ uses $\mathcal{V}_p^{(n)}$.

\definecolor{deepgray}{gray}{0.3}

\begin{table*}[t]
	\centering
    \vspace{-3mm}
    \caption{Comparison of SLT performance. SLT-FT denotes task-specific fine-tuning from the unified model.}
	\resizebox{1.0\textwidth}{!}{
	\begin{tabular}{l|cc|lll|lll|lll|lll}
		\toprule
		\multirow{3}{*}{SLT}
		& \multicolumn{2}{c|}{Extra}
		& \multicolumn{6}{c}{Phoenix14T}
		& \multicolumn{6}{c}{CSL-Daily} \\
		&&& \multicolumn{3}{c|}{DEV} & \multicolumn{3}{c}{TEST}
		& \multicolumn{3}{c|}{DEV} & \multicolumn{3}{c}{TEST} \\
		& Pose & RGB
		& ROUGE & BLEU1 & BLEU4
		& ROUGE & BLEU1 & BLEU4
		& ROUGE & BLEU1 & BLEU4
		& ROUGE & BLEU1 & BLEU4 \\
		\cline{1-15}

		\multicolumn{15}{l}{\textit{RGB-based}}\\
		\midrule

		\color{deepgray} SLRT~\cite{camgoz2020sign} &  & \color{deepgray} \cmark
		& \color{deepgray} -- & \color{deepgray} 47.26 & \color{deepgray} 22.38
		& \color{deepgray} -- & \color{deepgray} 46.61 & \color{deepgray} 21.32
		& \color{deepgray} 37.96 & \color{deepgray} 37.47 & \color{deepgray} 11.88
		& \color{deepgray} 36.74 & \color{deepgray} 37.38 & \color{deepgray} 11.79 \\

		\color{deepgray} STN-SLT~\cite{voskou2021stochastic} &  & \color{deepgray} \cmark
		& \color{deepgray} -- & \color{deepgray} 49.12 & \color{deepgray} 23.23
		& \color{deepgray} -- & \color{deepgray} 48.61 & \color{deepgray} 23.65
		& \color{deepgray} -- & \color{deepgray} -- & \color{deepgray} --
		& \color{deepgray} -- & \color{deepgray} -- & \color{deepgray} -- \\

		\color{deepgray} STMC-T~\cite{zhou2021spatial} &  & \color{deepgray} \cmark
		& \color{deepgray} 48.24 & \color{deepgray} 47.60 & \color{deepgray} 24.09
		& \color{deepgray} 46.65 & \color{deepgray} 46.98 & \color{deepgray} 23.65
		& \color{deepgray} -- & \color{deepgray} -- & \color{deepgray} --
		& \color{deepgray} -- & \color{deepgray} -- & \color{deepgray} -- \\

		\color{deepgray} SignBT~\cite{zhou2021improving} &  & \color{deepgray} \cmark
		& \color{deepgray} 50.29 & \color{deepgray} 51.11 & \color{deepgray} 24.45
		& \color{deepgray} 49.54 & \color{deepgray} 50.80 & \color{deepgray} 24.32
		& \color{deepgray} 49.49 & \color{deepgray} 51.46 & \color{deepgray} 20.80
		& \color{deepgray} 49.31 & \color{deepgray} 51.42 & \color{deepgray} 21.34 \\

		\color{deepgray} MMTLB~\cite{chen2022simple} &  & \color{deepgray} \cmark
		& \color{deepgray} 53.10 & \color{deepgray} 53.95 & \color{deepgray} 27.61
		& \color{deepgray} 52.65 & \color{deepgray} 53.97 & \color{deepgray} 28.39
		& \color{deepgray} 53.38 & \color{deepgray} 53.81 & \color{deepgray} 24.42
		& \color{deepgray} 53.25 & \color{deepgray} 53.31 & \color{deepgray} 23.92 \\

		\color{deepgray} BN-TIN-Transf.~\cite{zhou2021improving} &  & \color{deepgray} \cmark
		& \color{deepgray} -- & \color{deepgray} -- & \color{deepgray} --
		& \color{deepgray} -- & \color{deepgray} -- & \color{deepgray} --
		& \color{deepgray} 37.29 & \color{deepgray} 40.66 & \color{deepgray} 12.73
		& \color{deepgray} 37.67 & \color{deepgray} 40.74 & \color{deepgray} 13.19 \\

		\color{deepgray} COSLRT~\cite{gan2023contrastive} &  & \color{deepgray} \cmark
		& \color{deepgray} 52.47 & \color{deepgray} 52.29 & \color{deepgray} 27.83
		& \color{deepgray} 52.24 & \color{deepgray} 52.48 & \color{deepgray} 27.88
		& \color{deepgray} -- & \color{deepgray} -- & \color{deepgray} --
		& \color{deepgray} -- & \color{deepgray} -- & \color{deepgray} -- \\

		\color{deepgray} TwoStream-SLT~\cite{chen2022two} & \color{deepgray} \cmark & \color{deepgray} \cmark
		& \color{deepgray} 54.08 & \color{deepgray} 54.32 & \color{deepgray} 28.66
		& \color{deepgray} 53.48 & \color{deepgray} 54.90 & \color{deepgray} 28.95
		& \color{deepgray} 55.10 & \color{deepgray} 55.21 & \color{deepgray} 25.76
		& \color{deepgray} 55.72 & \color{deepgray} 55.44 & \color{deepgray} 25.79 \\

		\color{deepgray} SignDINO~\cite{gan2026learning} & \color{deepgray}  & \color{deepgray} \cmark
		& \color{deepgray} 53.61 & \color{deepgray} 53.49 & \color{deepgray} 27.17
		& \color{deepgray} 53.79 & \color{deepgray} 54.15 & \color{deepgray} 27.17
		& \color{deepgray} 52.36 & \color{deepgray} 53.64 & \color{deepgray} 25.62
		& \color{deepgray} 52.75 & \color{deepgray} 52.13 & \color{deepgray} 25.46 \\        

		\color{deepgray} MixSignGraph~\cite{gan2025mixsigngraph} & \color{deepgray}  & \color{deepgray} \cmark
		& \color{deepgray} 55.77 & \color{deepgray} 55.01 & \color{deepgray} 29.00
		& \color{deepgray} 53.84 & \color{deepgray} 54.90 & \color{deepgray} 28.97
		& \color{deepgray} 54.54 & \color{deepgray} 55.87 & \color{deepgray} 25.77
		& \color{deepgray} 54.67 & \color{deepgray} 55.41 & \color{deepgray} 25.87 \\

		\midrule
		\multicolumn{15}{l}{\textit{Pose-based}}\\
		\midrule
		Skeletor~\cite{jiang2021skeletor} & \cmark & 
		& 32.66 & 31.97 & 10.91
		& 31.80 & 31.86 & 10.35
		& -- & -- & --
		& -- & -- & -- \\
		
		Signbert+~\cite{hu2023signbert+} & \cmark & 
		& 45.53 & 44.45 & 19.86
		& 44.89 & 44.35 & 20.41
		& -- & -- & --
		& -- & -- & -- \\

		VAP~\cite{jiao2024visual} & \cmark & 
		& \textbf{51.47} & \textbf{52.78} & \textbf{26.62}
		& \textbf{51.28} & \textbf{53.07} & \textbf{26.16}
		& 48.72 & 50.41 & 21.16
		& 48.56 & 49.99 & 20.85 \\

        \rowcolor{LightBlue}
		Uni-SLTP & \cmark & 
		& 50.92 & 51.99 & 25.00
		& 50.47 & 52.29 & 25.75
		& \textbf{50.11} & \textbf{51.27} & \textbf{23.06}
		& \textbf{49.82} & \textbf{51.96} & \textbf{23.63} \\

        \rowcolor{LightBlue}
		Uni-SLTP (SLT-FT) & \cmark & 
		& \textbf{51.88} & \textbf{52.93} & \textbf{26.94}
		& \textbf{52.39} & \textbf{53.25} & \textbf{26.89}
		& \textbf{50.73} & \textbf{51.41} & \textbf{23.40}
		& \textbf{50.24} & \textbf{52.27} & \textbf{23.86} \\

		\bottomrule
	\end{tabular}}
	\label{tab:SLT}
\end{table*}
\begin{table*}[t]
\centering 
\caption{Comparison of SLP performance. * denotes reimplemented results; SLP-FT denotes fine-tuning from the unified model.}
\resizebox{1.0\textwidth}{!}{
\small
\setlength{\tabcolsep}{4pt}
\renewcommand{\arraystretch}{1.12}
\begin{tabular}{l|cccc|cccc|cccc|cccc}
\toprule
\multirow{3}{*}{SLP}
& \multicolumn{8}{c|}{Phoenix14T}
& \multicolumn{8}{c}{CSL-Daily} \\
& \multicolumn{4}{c|}{DEV} & \multicolumn{4}{c|}{TEST}
& \multicolumn{4}{c|}{DEV} & \multicolumn{4}{c}{TEST} \\
& \multicolumn{2}{c}{{B-T}$\uparrow$} & \multicolumn{2}{c|}{{DTW-PA-JPE}$\downarrow$}
& \multicolumn{2}{c}{{B-T}$\uparrow$} & \multicolumn{2}{c|}{{DTW-PA-JPE}$\downarrow$}
& \multicolumn{2}{c}{{B-T}$\uparrow$} & \multicolumn{2}{c|}{{DTW-PA-JPE}$\downarrow$}
& \multicolumn{2}{c}{{B-T}$\uparrow$} & \multicolumn{2}{c}{{DTW-PA-JPE}$\downarrow$} \\

\cmidrule(lr){2-3}\cmidrule(lr){4-5}
\cmidrule(lr){6-7}\cmidrule(lr){8-9}
\cmidrule(lr){10-11}\cmidrule(lr){12-13}
\cmidrule(lr){14-15}\cmidrule(lr){16-17}

& ROUGE & BLEU4 & Body & Hand
& ROUGE & BLEU4 & Body & Hand
& ROUGE & BLEU4 & Body & Hand
& ROUGE & BLEU4 & Body & Hand \\
\midrule
\multicolumn{17}{l}{\textit{Text2Gloss2Pose}}\\
\midrule
PT~\cite{saunders2020progressive}  
& 11.87 & 3.88 & 14.33 & 10.47
& 13.17 & 4.31 & 13.15 & 10.26
& 7.68 & 0.67 & 15.85  & 13.97 
& 7.54 & 0.41 & 16.32  & 13.29 \\
% G2P-DDM~\cite{xie2024g2p}  
% & 20.05 & 6.36 & 10.91 & 7.55
% & 20.37 & 6.25 & 10.36 & 7.07
% & 8.25 & 3.12 &   &  
% & 8.11 & 3.86 &   &   \\
Sign-IDD~\cite{tang2025sign}  
& 27.97 & 8.42 & 9.16 & 3.07
& 27.11 & 8.46 & 9.00 & 3.05
& 14.70 & 2.98 & 13.13  & 4.52 
& 14.11 & 2.74 & 13.24  & 4.69  \\
\midrule
\multicolumn{17}{l}{\textit{Text2Pose}}\\
\midrule
SignPR~\cite{liu2026signpr}  
& 30.84 & 9.12 & 6.14 & 1.90
& 32.86 & 9.41 & 6.08 & 1.87
& 15.02 & 3.36 & 12.97  & 4.21 
& 14.43 & 3.01 & 12.82  & 4.56  \\

T2M-GPT*~\cite{zhang2023generating}  
& 28.53 & 8.40 & 8.95 & 2.95
& 28.97 & 8.41 & 8.76 & 2.60
& 25.12 & 6.71 & 8.95 & 2.34
& 26.57 & 6.93 & 8.44 & 2.32 \\

SOKE*~\cite{zuo2025signs} 
& 30.94 & 10.15 & 6.39 & 1.91
& 31.23 & 10.43 & 6.16 & 1.85
& 27.85 & 9.74 & 7.81 & 2.05
& 28.29 & 10.31 & 7.58 & 2.17 \\

\rowcolor{LightBlue}
Uni-SLTP 
& \textbf{33.41} & \textbf{11.04} & \textbf{5.94} & \textbf{1.76}
& \textbf{34.27} & \textbf{11.94} & \textbf{5.80} & \textbf{1.72}
& \textbf{29.74} & \textbf{11.53} & \textbf{6.99} & \textbf{1.86}
& \textbf{30.02} & \textbf{11.41} & \textbf{6.32} & \textbf{1.71} \\

\rowcolor{LightBlue}
Uni-SLTP (SLP-FT)  
& \textbf{34.09} & \textbf{11.76} & \textbf{5.49} & \textbf{1.69}
& \textbf{35.73} & \textbf{12.04} & \textbf{5.54} & \textbf{1.67}
& \textbf{30.18} & \textbf{11.95} & \textbf{6.67} & \textbf{1.73}
& \textbf{30.36} & \textbf{12.01} & \textbf{6.28} & \textbf{1.65} \\

\bottomrule
\end{tabular}
}
\label{tab:SLP}
\vspace{-1mm}
\end{table*}

\paragraph{Training and inference.}
Uni-SLTP is trained in two stages.
In Stage~1, we pretrain the sign tokenizer $\mathbf{ST}$ with $\mathcal{L}_{\text{tok}}$ (Eq.~\ref{eq:loss_tok}) and then freeze $\mathbf{ST}_e$ and $\mathbf{ST}_d$.
In Stage~2, we fine-tune the pretrained seq2seq LM as the autoregressive module $\mathbf{AR}$ under the unified vocabulary $\mathcal{V}$. Rather than directly mixing SLT and SLP from scratch, we adopt a progressive training schedule: we first optimize the model on SLP, then on SLT, and finally perform joint training on both tasks.

\textit{SLT.}
In training, given pose--text pairs $(\mathbf{S},\mathbf{W})$, we tokenize the pose input with the frozen $\mathbf{ST}_e$ and use only the semantic-stage IDs $\mathbf{Q}_1$ as the source sequence and the text tokens $\mathbf{W}^{id}$ as the target.
We train with standard teacher-forced negative log-likelihood:
\begin{equation}
\label{eq:loss_slt}
% \vspace{-1mm}
\mathcal{L}_{\text{SLT}}
= - \sum_{u=1}^{U} \log p_\theta\!\left(w_u^{id}\mid \mathbf{W}^{id}_{<u}, \mathbf{Q}_1\right).
\vspace{-1mm}
\end{equation}
At inference, $\mathbf{AR}$ autoregressively generates tokens conditioned on $\mathbf{Q}_1$, and the LM decoder converts predicted IDs back to the sentence.

\textit{SLP.}
In training, given text--pose pairs $(\mathbf{W},\mathbf{S})$, we encode text with the LM and tokenize the target pose with the frozen $\mathbf{ST}_e$ to obtain RVQ indices $\{\mathbf{Q}_n\}_{n=1}^{N}$.
We serialize the stage-wise indices into a single pose-token sequence
$\mathbf{P}^{id} = [\texttt{<sos>}, q_{1,1},\ldots,q_{N,1}, \ldots, \\
q_{1,T'},\ldots,q_{N,T'}, \texttt{<eos>}]$,
where each $q_{n,t}$ is mapped to its stage-specific token ID in $\mathcal{V}^{(n)}_p$.
We optimize the pose-token likelihood with:
\begin{equation}
\label{eq:loss_slp}
% \vspace{-1mm}
\mathcal{L}_{\text{SLP}}
= - \sum_{\ell=1}^{L} \log p_\theta\!\left(p_\ell^{id}\mid \mathbf{P}^{id}_{<\ell}, \mathbf{W}\right).
% \vspace{-1mm}
\end{equation}
At inference, $\mathbf{AR}$ generates pose tokens until \texttt{<eos>}; we then de-serialize them back into $\{\hat{\mathbf{Q}}_n\}_{n=1}^{N}$ and reconstruct poses with the frozen decoder $\mathbf{ST}_d$.

\section{Experiments}
\begin{table}[t]
  \centering
    \caption{Comparison with unified human motion methods on SLT and SLP tasks on the Phoenix14T dataset.}
  \setlength{\tabcolsep}{4pt}
  \renewcommand{\arraystretch}{1.15}
  \resizebox{1.0\linewidth}{!}{
  \begin{tabular}{l|cc|ccc}
    \toprule
    \multirow{2}{*}{Methods}
    & \multicolumn{2}{c|}{SLT Task}
    & \multicolumn{3}{c}{SLP Task} \\
    \cmidrule(lr){2-3}\cmidrule(lr){4-6}
    & ROUGE$\uparrow$ & BLEU4$\uparrow$
    & BLEU4$\uparrow$ & DTW-Body$\downarrow$ & DTW-Hand$\downarrow$ \\
    \midrule
    MotionGPT~\cite{jiang2023motiongpt} & 35.36 & 10.04 & 6.06 & 10.64 & 6.89 \\
    MotionGPT2~\cite{wang2024motiongpt} & 37.96 & 11.88 & 8.62 & 9.34 & 3.41 \\
    MG-MotionLLM~\cite{wu2025mg}  & 36.67 & 11.39 & 8.46 & 9.78 & 3.67 \\
    \rowcolor{LightBlue}
    Uni-SLTP & \textbf{50.47} & \textbf{25.75} &\textbf{11.94} &\textbf{5.80} &\textbf{1.72}  \\
    
    \bottomrule
  \end{tabular}
  }
  \label{tab:motion}
  \vspace{-5mm}
\end{table}

\subsection{Experimental Setup}
\paragraph{Datasets.} We evaluate our unified framework for SLT and SLP on two widely used datasets: Phoenix14T~\cite{camgoz2018neural} and CSL-Daily~\cite{zhou2021improving}. Phoenix14T, a German Sign Language dataset from weather forecasts, contains 8,257 videos (7,096/519/642 train/dev/test) across 9 signers, with 1,066 glosses and 2,887 German words. CSL-Daily, a Chinese Sign Language dataset covering daily-life topics, contains 20,654 videos (18,401/1,077/1,176 train/dev/test) from 10 signers, with 2,000 glosses and 2,343 Chinese words.  Following SOKE~\cite{zuo2025signs}, we represent each SL motion sequence as $\mathbf{S}\in\mathbb{R}^{T\times d}$, where $T$ is the sequence length and $d=133$ denotes the number of SMPL-X parameters, including 11 upper-body joints, 30 hand joints, and 10 expression parameters.

\iffalse
We evaluate our unified framework for SLT and SLP on two widely used datasets: RWTH-PHOENIX-Weather 2014T (Phoenix14T) \cite{camgoz2018neural} and CSL-Daily \cite{zhou2021improving}. Phoenix14T is a German Sign Language dataset collected from weather forecasts, containing 8,257 videos split into 7,096/519/642 for train/dev/test across 9 signers. It provides both gloss and spoken-language annotations, with a gloss vocabulary of 1,066 unique signs and a German translation vocabulary of 2,887 words. CSL-Daily is a Chinese Sign Language dataset covering daily-life topics, consisting of 20,654 videos from 10 signers, split into 18,401/1,077/1,176 for train/dev/test. It includes 2,000 gloss categories and a Chinese translation vocabulary of 2,343 words. Following SOKE~\cite{zuo2025signs}, we represent each SL motion sequence as $\mathbf{S}\in\mathbb{R}^{T\times d}$, where $T$ is the sequence length and $d=133$ denotes the number of SMPL-X parameters, including 11 upper-body joints, 30 hand joints, and 10 expression parameters.
\fi

\paragraph{Implementation Details.}
% (1) \textit{SR-RVQ.} We tokenize continuous pose sequences using a 3-level RVQ module. The pose encoder applies a temporal downsampling factor of $4$, producing latent features of length $T' = T/4$. We utilize three-stage RVQ codebooks $\{Q_1, Q_2, Q_3\}$, each containing $512$ entries. Following our semantic-detail factorization, $Q_1$ is regularized by semantic supervision (via CTC and contrastive objectives) to capture coarse semantic structures, while $Q_2$ and $Q_3$ model residual motion details to enhance reconstruction accuracy.
% (2) \noindent\textit{Autoregressive Model.} For the sequence modeling, we adopt \textsc{mBART-large-cc25}~\cite{liu2020multilingual} as the backbone, which consists of 12 encoder-decoder layers with a model dimension of 1024. More training details are provided in the Appendix.
(1) \textit{SR-RVQ.} We tokenize continuous pose sequences using a 3-level RVQ module. 
The pose encoder downsamples the temporal length by a factor of 4, yielding latent features of length $T' = T/4$. 
The RVQ module comprises three codebooks, $\{Q_1, Q_2, Q_3\}$, each containing 512 entries. 
Under our semantic-reconstruction decomposition, $Q_1$ is regularized by CTC and contrastive supervision to encode coarse semantic structures, while $Q_2$ and $Q_3$ capture residual motion details for more accurate reconstruction.
(2) \noindent\textit{Autoregressive Model.} We adopt \textsc{mBART-large-cc25}~\cite{liu2020multilingual} as the backbone for sequence modeling. 
It consists of 12 encoder-decoder layers with a hidden size of 1024. 
% \textit{More training details are provided in the Appendix.}

\paragraph{Evaluation Metrics.}
For SLT, we report ROUGE-L~\cite{lin2004automatic} and BLEU-1/BLEU-4~\cite{papineni2002bleu} on the generated sentences.
For SLP, we evaluate both semantic consistency and motion accuracy.
(1) \emph{Back-translation (B-T):} We translate the generated poses back to text by employing a translation model~\cite{gan2025mixsigngraph}, and report ROUGE-L and BLEU-1/4 scores against the ground truth.
(2) \emph{Motion metrics:} following SOKE~\cite{zuo2025signs}, to handle length mismatch between generated and reference signs, we report JPE/MPJPE in the original and Procrustes-aligned spaces, and their DTW counterparts (DTW-PA-JPE), where DTW temporally aligns the generated and reference pose sequences while computing the joint position error.

\begin{figure*}[t]
  \centering
  \includegraphics[width=1.0\linewidth]{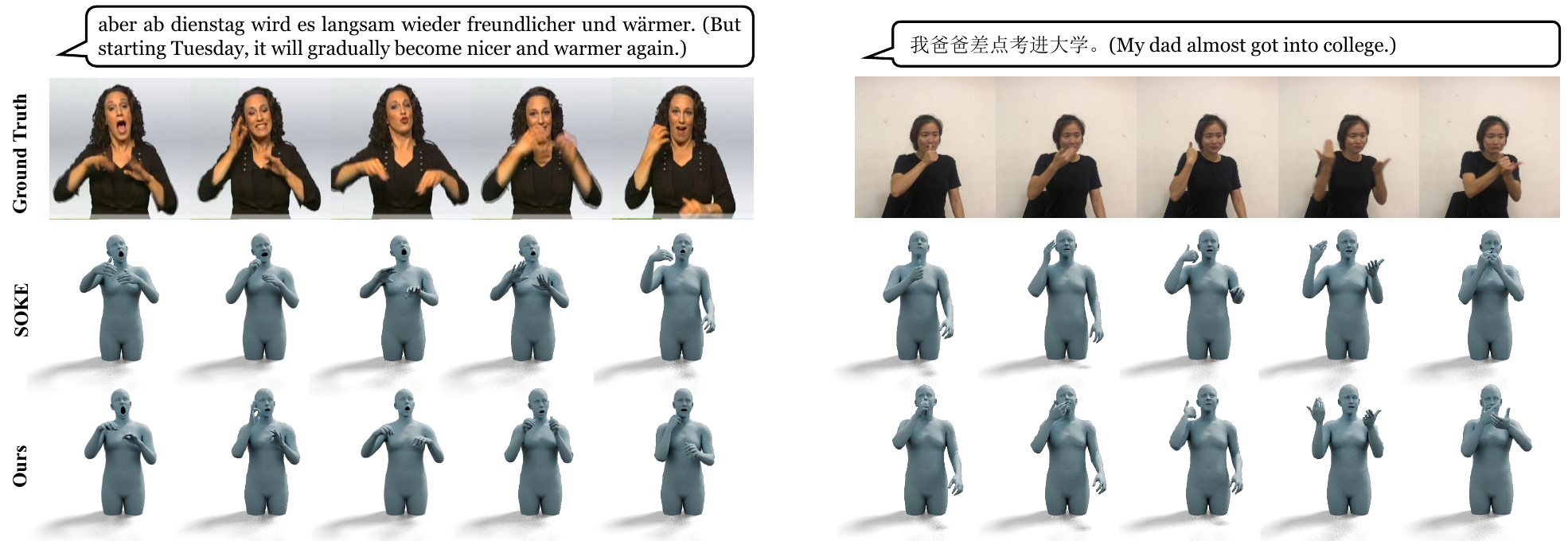}
  \vspace{-4mm}
  \caption{SLP qualitative results of our Uni-SLTP and baseline method SOKE on Phoenix14T (left) and CSL-Daily (right) datasets.}
  \label{fig:aba}
  \vspace{-1mm}
\end{figure*}

\begin{table*}[t]
\centering
\caption{Ablation of the SR-RVQ tokenizer. PA-MPJPE measures reconstruction quality; WER is reported when CTC loss is used.}
\resizebox{1.0\textwidth}{!}{%
\begin{tabular}{l|c|ccc|c|cc|cc}
\toprule
\multirow{2}{*}{\textbf{Tokenizer Setting}}
& \textbf{RVQ}
& \multicolumn{3}{c|}{\textbf{Supervision}}
& \textbf{CSLR}
& \multicolumn{2}{c|}{\textbf{SLT}}
& \multicolumn{2}{c}{\textbf{Reconstruction}} \\
\cmidrule(lr){3-5}\cmidrule(lr){6-6}\cmidrule(lr){7-8}\cmidrule(lr){9-10}
& (Residuals)
& Recon
& CTC
& CL
& WER $\downarrow$
& ROUGE $\uparrow$
& BLEU-4 $\uparrow$
& PA-MPJPE-body $\downarrow$
& PA-MPJPE-hand $\downarrow$ \\
\midrule
{(a) VQ-VAE (Recon)}                  &  & \cmark &  &  &  -   & 19.71 & 4.08  & \textbf{13.02} & \textbf{4.38} \\
{(b) Semantic-VQ (CTC+Con)}            &  &  & \cmark & \cmark & \textbf{27.65} & \textbf{51.68} & \textbf{26.76} & 44.52 & 14.94 \\
{(c) VQ-VAE (Recon+CTC+Con)}           &  & \cmark & \cmark & \cmark & 44.61 & 36.55 & 13.97 & 16.73 & 6.84 \\
\midrule
{(d) SR-RVQ (w/o CTC)}                & \cmark & \cmark &  & \cmark &  -   & 21.24 & 4.65 & 13.19 & 4.48 \\
{(e) SR-RVQ (w/o Con)}                 & \cmark & \cmark & \cmark &  & 36.76 & 48.47 & 22.90 & 13.30 & 4.46 \\
\rowcolor{LightBlue}
{(f) SR-RVQ (Ours)}                   & \cmark & \cmark & \cmark & \cmark & 28.63 & 50.47 & 25.75 & 13.68 & 4.69 \\
\bottomrule
\end{tabular}}
\label{tab:rvq_design}
\vspace{-5pt}
\end{table*}

% \begin{table*}[t]
% \centering
% \caption{Effect of SR-RVQ stages on SLT and SLP.}
% \label{tab:rvq_stage}
% \resizebox{0.7\textwidth}{!}{%
% \begin{tabular}{l|cc|ccc|c}
% \toprule
% \multirow{2}{*}{RVQ stages used}
%     & \multicolumn{2}{c|}{SLT Task}
%     & \multicolumn{3}{c|}{SLP Task}
%     & Efficiency \\
%     & ROUGE $\uparrow$
%     & BLEU-4 $\uparrow$
%     & BLEU-4 $\uparrow$
%     & DTW-Body $\downarrow$
%     & DTW-Hand $\downarrow$
%     & s/video $\downarrow$ \\
% \midrule
% \textit{(a) $Q_1$}           & \textbf{50.47} & \textbf{25.75} & 0.93 & 18.76 & 15.60 & 1.06 \\
% \textit{(b) $Q_2 + Q_3$ } & 18.20 & 3.98 & 11.62 & 5.99 & 1.65 &  1.47    \\
% \textit{(c) $Q_1 + Q_2 + Q_3$}        & 45.78 & 19.23 & \textbf{12.94} & \textbf{5.80} & \textbf{1.72} &  1.68    \\
% \bottomrule
% \end{tabular}%
% }
% \vspace{-10pt}
% \end{table*}

\begin{table}[t]
\centering
\caption{Effect of SR-RVQ stages on SLT and SLP.}
\label{tab:rvq_stage}
\resizebox{1.0\columnwidth}{!}{%
\begin{tabular}{l|cc|cc|c}
\toprule
\multirow{2}{*}{RVQ stages used}
    & \multicolumn{2}{c|}{SLT Task}
    & \multicolumn{2}{c|}{SLP Task}
    & Efficiency \\
    & ROUGE $\uparrow$
    & BLEU-4 $\uparrow$
    & DTW-Body $\downarrow$
    & DTW-Hand $\downarrow$
    & s/video $\downarrow$ \\
\midrule
{(a) $\{Q_1\}$}            & \textbf{51.68} & \textbf{26.76} & 18.76 & 15.60 & 1.06 \\
{(b) $\{Q_2,Q_3\}$}        & 18.20 & 3.98 & 5.99 & 1.76 & 1.47 \\
{(c) $\{Q_1,Q_2,Q_3\}$}    & 45.78 & 19.23 & \textbf{5.80} & \textbf{1.72} & 1.68 \\
\bottomrule
\end{tabular}%
}
\vspace{-10pt}
\end{table}

\subsection{Comparisons}
% Since the back-translation model widely adopted in prior SLP work~\cite{saunders2020progressive,xie2024g2p,yin2024t2s} relies on an SLT model~\cite{camgoz2020sign} whose checkpoints are not publicly released, 
Following~\cite{liu2026signpr}, we train the SLT model~\cite{gan2025mixsigngraph} and use it to evaluate our generated signs.
% To make the comparison fair given the sensitivity to evaluator weights, we reproduce open-source SLP baselines (\ie PT, G2P-DDM, Sign-IDD, SOKE) under the same preprocessing and evaluation pipeline, and exclude methods without public implementations from our tables.

\paragraph{SLT Comparisons.}
As shown in Table~\ref{tab:SLT}, we evaluate Uni-SLTP on Phoenix14T and CSL-Daily against both RGB-based and pose-based SLT methods. Since Uni-SLTP is \emph{pose-only}, it is expected to underperform RGB-based models that exploit richer visual cues for SLT, especially compared to MixSignGraph. Nevertheless, under the same pose-based setting, Uni-SLTP remains competitive: it achieves performance comparable to the baseline VAP on Phoenix14T and yields an improvement on CSL-Daily. We further evaluate a task-specific SLT fine-tuned variant initialized from the unified model. This further improves SLT performance, indicating that the unified model already provides a strong shared foundation across tasks, while additional task-specific adaptation can further specialize the model for SLT. Overall, these results show that Uni-SLTP remains competitive among pose-based SLT methods, while also providing a unified framework that also supports SLP. 
% We also report an independently trained SLT-only variant, whose performance is slightly higher than that of the unified model on the SLT task. We consider this a reasonable trade-off for unified modeling, where a shared representation must simultaneously support both understanding and generation.

\paragraph{SLP Comparisons.}
Table~\ref{tab:SLP} reports the performance comparison between Uni-SLTP and prior SLP methods on the Phoenix14T and CSL-Daily datasets. Experimental results show that Uni-SLTP outperforms prior methods in both the B-T metric for semantic consistency and the motion-related metrics for motion accuracy. Task-specific fine-tuning for SLP further improves performance.

\paragraph{Comparisons with Unified Human Motion Methods.}
As shown in Table~\ref{tab:motion}, we compare Uni-SLTP with representative unified human motion methods on the Phoenix14T dataset. Directly applying these models to sign language tasks results in worse SLT performance and lower SLP performance than our method. One possible reason is that the alignment between sign sequences and text is more semantically complex and often non-monotonic, making it difficult for methods designed for generic human motion to model the fine-grained linguistic structure of sign language effectively.

\subsection{Qualitative Results}
We provide a qualitative comparison between our Uni-SLTP and the baseline method SOKE on the SLP task. As shown in Figure~\ref{fig:aba}, the sign sequences generated by Uni-SLTP exhibit finer hand-level details that are closer to the ground truth, resulting in lower motion error and more natural signing dynamics.

\begin{table}[t]
  \centering
    \caption{Effect of the choice of the pre-trained LM backbone.}
  \setlength{\tabcolsep}{4pt}
  \renewcommand{\arraystretch}{1.15}
  \resizebox{1.0\linewidth}{!}{
  \begin{tabular}{l|cc|ccc}
    \toprule
    \multirow{2}{*}{Backbone}
    & \multicolumn{2}{c|}{SLT Task}
    & \multicolumn{3}{c}{SLP Task} \\
    \cmidrule(lr){2-3}\cmidrule(lr){4-6}
    & ROUGE$\uparrow$ & BLEU4$\uparrow$
    & BLEU4$\uparrow$ & DTW-Body$\downarrow$ & DTW-Hand$\downarrow$ \\
    \midrule
    Llama3.2 1B~\cite{touvron2023llama} & 49.61 & 25.33 & 11.58 & 6.02 & 1.75 \\
    Gemma~\cite{gemma_2025} & 49.99 & 25.36 & 11.32 & 5.97 & 1.76 \\
    mT5~\cite{xue2021mt5}  & 50.99 & 25.36 & 11.97 & 5.78 & 1.70 \\
    \rowcolor{LightBlue}
    mBART (ours) & \textbf{50.47} & \textbf{25.75} &\textbf{11.94} &\textbf{5.80} &\textbf{1.72}  \\
    
    \bottomrule
  \end{tabular}
  }
  \label{tab:llm}
  \vspace{-5mm}
\end{table}

\begin{table}[t]
  \centering
  \caption{Ablation study of different loss weight combinations.}
  \setlength{\tabcolsep}{4pt}
  \renewcommand{\arraystretch}{1.15}
  \resizebox{1.0\linewidth}{!}{
  \begin{tabular}{ccc|cc|cc}
    \toprule
    \multicolumn{3}{c|}{Loss Weights}
    & \multicolumn{2}{c|}{SLT Task}
    & \multicolumn{2}{c}{Reconstruction} \\
    \cmidrule(lr){1-3}\cmidrule(lr){4-5}\cmidrule(lr){6-7}
    $\lambda_{\mathrm{rec}}$ & $\lambda_{\mathrm{CTC}}$ & $\lambda_{\mathrm{Con}}$
    & ROUGE$\uparrow$ & BLEU-4$\uparrow$
    & PA-MPJPE-body$\downarrow$ & PA-MPJPE-hand$\downarrow$ \\
    \midrule
    1.0 & 0   & 1.0 & 21.24 & 4.65  & \textbf{13.19} & 4.48 \\
    1.0 & 1.0 & 0   & 48.47 & 22.90 & 13.30          & \textbf{4.46} \\
    1.0 & 1.0 & 1.0 & 46.29 & 23.74 & 15.59          & 5.56 \\
    1.0 & 0.2 & 0.8 & 25.28 & 7.62  & 13.24          & 4.52 \\
    \rowcolor{LightBlue}
    1.0 & 0.8 & 0.2 & \textbf{50.47} & \textbf{25.75} & 13.68 & 4.69 \\
    \bottomrule
  \end{tabular}
  }
  \label{tab:loss}
  \vspace{-2mm}
\end{table}

\begin{table}[t]
  \centering
  \caption{Ablation on RVQ depth.
  \emph{Eff.\ \#Codes} ($\uparrow$) denotes the effective number of utilized codes.
  % (computed as $\exp(H(p(q)))$ from code usage and averaged over stages).
  }
  \vspace{-3pt}
  \setlength{\tabcolsep}{4pt}
  \renewcommand{\arraystretch}{1.15}
  \resizebox{1.0\linewidth}{!}{
  \begin{tabular}{c|c|cc|ccc}
    \toprule
    \multirow{2}{*}{RVQ Depth $N$}
    & \multirow{2}{*}{Eff.\ \#Codes $\uparrow$}
    & \multicolumn{2}{c|}{SLT Task}
    & \multicolumn{3}{c}{SLP Task} \\
    \cmidrule(lr){3-4}\cmidrule(lr){5-7}
    & 
    & ROUGE$\uparrow$ & BLEU4$\uparrow$
    & BLEU4$\uparrow$ & DTW-Body$\downarrow$ & DTW-Hand$\downarrow$ \\
    \midrule
    1 & 385   & \textbf{36.55}& \textbf{13.97} & 8.64  & 7.71 & 2.06 \\
    % 1 & 385   & \textbf{52.31} & \textbf{27.09} & 1.84  & 43.75 & 15.00 \\
    2 & 416   & 50.91 & 25.00 & 10.67 & 7.29 & 1.87 \\
    \rowcolor{LightBlue}
    3 & 338 & 50.47 & 25.75 & \textbf{11.94} & \textbf{5.80} & \textbf{1.72} \\
    4 & 280   & 50.63 & 25.54 & 11.69 & 5.82 & 1.73 \\
    \bottomrule
  \end{tabular}
  }
  \label{tab:depth}
  \vspace{-2mm}
\end{table}

\begin{table}[t]
  \centering
    \caption{Ablation on codebook size.}
  \setlength{\tabcolsep}{4pt}
  \renewcommand{\arraystretch}{1.15}
  \resizebox{1.0\linewidth}{!}{
  \begin{tabular}{c|cc|ccc}
    \toprule
    \multirow{2}{*}{Codebook Size $K$}
    & \multicolumn{2}{c|}{SLT Task}
    & \multicolumn{3}{c}{SLP Task} \\
    \cmidrule(lr){2-3}\cmidrule(lr){4-6}
    & ROUGE$\uparrow$ & BLEU4$\uparrow$
    & BLEU4$\uparrow$ & DTW-Body$\downarrow$ & DTW-Hand$\downarrow$ \\
    \midrule
    128 & 46.40 & 22.87 & 11.68 & 5.70 & 1.74 \\
    256 & 49.82 & 25.64 & 11.62 & 5.68 & 1.71 \\
    \rowcolor{LightBlue}
    512 & 50.47 & 25.75 &\textbf{11.94} &\textbf{5.80} &\textbf{1.72}  \\
    1024  & \textbf{50.94} & \textbf{26.11} & 11.05 & 6.07 & 1.83 \\
    \bottomrule
  \end{tabular}
  }
  \label{tab:size}
  % \vspace{-2mm}
\end{table}

\subsection{Ablation Study}
All ablation experiments were conducted on the Phoenix14T dataset.
\paragraph{Effect of SR-RVQ Tokenizer.}
To assess the impact of tokenizer design on SLT and reconstruction, we compare representative variants in Table~\ref{tab:rvq_design}.
\textit{(a) VQ-VAE} (reconstruction-only) achieves the best PA-MPJPE but poor language metrics, while \textit{(b) Semantic-VQ} (CTC+Con) substantially improves CSLR/SLT yet severely degrades reconstruction. \textit{(c) VQ-VAE (Recon+CTC+Con)} adds semantic loss to a single-bottleneck VQ-VAE but still fails to balance semantics and motion accuracy. 
Turning to our SR-RVQ variants, the results with residual codebooks enabled show that both semantic signals matter:
% Enabling residual codebooks, SR-RVQ variants show that both semantic signals matter:
removing CTC (\textit{d}) notably hurts SLT, and removing contrastive learning (\textit{e}) reduces CSLR/SLT. \textit{(f) SR-RVQ (Ours)} combines residual quantization with both CTC and contrastive supervision, yielding the best overall trade-off, with near semantic-only performance and reconstruction close to the VQ-VAE baseline.

% \paragraph{Effect of Autoregressive Modeling Strategy.}
% With $N{=}3$, we compare our $\mathbf{AR}$+$\mathbf{DT}$ decoding with a \textit{Flattened AR} baseline that predicts all RVQ codes as one sequence. Flattened AR increases the effective length to $3T'$, while ours keeps it at $T'$. In Table~\ref{tab:slp_decode}, our method is $\sim$40\% faster and achieves lower DTW on both body and hand, indicating reduced error accumulation.
\paragraph{Effect of SR-RVQ Stages.}
Table~\ref{tab:rvq_stage} examines how using different SR-RVQ stages at training and inference affects SLT, SLP, and efficiency.
Using (a) only the semantic stage gives the best SLT results but poor SLP motion quality, while using (b) only residual stages improves SLP but severely degrades SLT. 
% With (c) all stages yields the best overall trade-off, achieving the strongest SLP metrics while incurring an expected but acceptable runtime overhead due to generating additional residual tokens. 
Using (c) all stages yields the best overall trade-off, achieving competitive SLP metrics with acceptable runtime overhead. Therefore, we use the semantic stage for SLT and all stages for SLP.

% \begin{figure}[t]
%   \centering
%   \includegraphics[width=0.97\columnwidth]{sec/figure/user.pdf}
%   \vspace{-2mm}
%   \caption{User study for SLP.}
% \label{fig:user1}
%   \vspace{-5mm}
% \end{figure}

\paragraph{Effect of the Choice of the Pre-trained LM Backbone.}
To study the impact of the pre-trained LM, we replace mBART with other open-source language backbones, including Llama3.2 1B, Gemma, and mT5. As shown in Table~\ref{tab:llm}, different backbones lead to consistent performance with only minor variations, and no single model substantially changes task balance. This suggests that our framework is not sensitive to the choice of pre-trained LM, and the improvements primarily come from our unified sign representation and training design rather than the backbone itself.

\paragraph{Effect of Loss Weights in the SR-RVQ Tokenizer.}
To evaluate the effect of the loss weights in SR-RVQ on SLT and reconstruction, we conduct a sensitivity analysis in Table~\ref{tab:loss}. The results show that increasing the semantic-loss weights improves SLT performance but degrades reconstruction quality, while larger reconstruction weights lead to the opposite trend. In the experiments, we choose the setting $(\lambda_{\mathrm{rec}}=1.0,\lambda_{\mathrm{CTC}}=0.8,\lambda_{\mathrm{Con}}=0.2)$.

\paragraph{Effect of RVQ Depth.}
To study RVQ depth, Table~\ref{tab:depth} varies the number of stages $N$ and reports both downstream metrics and code usage.  A single-stage VQ ($N{=}1$), where semantic and reconstruction objectives share the same codebook, underperforms on SLT/SLP despite relatively high code usage, while increasing depth markedly improves performance. $N{=}3$ gives the best overall trade-off, achieving the strongest SLP quality with balanced code usage; moving to $N{=}4$ yields only marginal gains, indicating diminishing returns. Accordingly, we adopt a three-stage SR-RVQ in all experiments.

\paragraph{Effect of Codebook Size.}
Codebook size controls the representational capacity of each RVQ stage. We therefore fix the RVQ depth to $N{=}3$ and vary the codebook size $K$ to evaluate both SLT and SLP (Table~\ref{tab:size}). A small codebook ($K{=}128$) underperforms on both tasks, indicating limited expressiveness. In contrast, an overly large codebook ($K{=}1024$) yields only marginal SLT gains but harms SLP, with lower BLEU-4 and higher DTW errors. Overall, $K{=}512$ provides the best trade-off between semantic modeling and motion detail recovery, achieving the strongest overall performance.

% \paragraph{User Study.}
% To mitigate potential bias of back-translation based SLP evaluation, we conduct a human study with 10 CSL-trained participants.
% Each participant rates 15 samples on Phoenix14T and 15 on CSL-Daily using a 1--10 Likert scale (higher is better), with anonymized method labels and randomized presentation order.
% As shown in Figure~\ref{fig:user1}, Uni-SLTP achieves higher average scores on all criteria; detailed settings are provided in Appendix.

\paragraph{Model Size and Training/Inference Speed.} As shown in Table~\ref{tab:efficiency}, we report the trainable parameters, training and inference efficiency of our model on Phoenix14T using a single NVIDIA A6000 GPU. Training time is measured on the training set, while inference speed is evaluated on test samples with an average sequence length of 250 frames, averaged over 100 runs. Overall, Uni-SLTP enables efficient unified training, while SLP inference is slower due to autoregressive pose generation.

\begin{table}[t]
  \centering
  \caption{Efficiency statistics of Uni-SLTP.}
  \setlength{\tabcolsep}{5pt}
  \renewcommand{\arraystretch}{1.12}
  \resizebox{1.0\linewidth}{!}{
  \begin{tabular}{l|c|c|c}
    \toprule
    Phase & Model & Trainable \#Params (M) & Efficiency \\
    \midrule
    Tokenizer training & SR-RVQ & 76.95 & 4 min/epoch \\
    Unified task training & AR model & 375.88 & 15 min/epoch \\
    SLT inference & Uni-SLTP & N/A & 12.6 samples/s \\
    SLP inference & Uni-SLTP & N/A & 0.60 samples/s \\
    \bottomrule
  \end{tabular}}
  \label{tab:efficiency}
  \vspace{-4mm}
\end{table}
 
\section{Conclusion}
We revisit SLT and SLP as two inverse directions of sign and text mapping, and show that unifying them is feasible only if the sign representation supports both semantic alignment for SLT and motion accuracy for SLP. We propose Uni-SLTP with SR-RVQ, a hierarchical pose tokenizer that converts continuous pose sequences into discrete multi-stage tokens. It separates a semantic anchor token from residual motion detail tokens to bridge the Semantic-Reconstruction Gap.
With the resulting shared token space, both tasks reduce to conditional next token prediction and can be handled by a single language model. Experiments on two widely used datasets show stronger SLP performance while keeping pose-based SLT competitive with task-specific baselines.

\begin{acks}
This work is supported in part by National Natural Science Foundation of China under Grant Nos. 62172208, 92467202, 62272216; Key Projects of Jiangsu Provincial Basic Research Program under Grant No. BK20243040; JiangSu Natural Science Foundation under Grant No. BK20251989. This work is partially supported by Fundamental and Interdisciplinary Disciplines Breakthrough Plan of the Ministry of Education of China (No. JYB2025XDXM118); the “111 Center” (No. B26023); Collaborative Innovation Center of Novel Software Technology and Industrialization.
\end{acks}

\bibliographystyle{ACM-Reference-Format}
\balance
\bibliography{bib/new, bib/newother, bib/newslt}

\end{document}